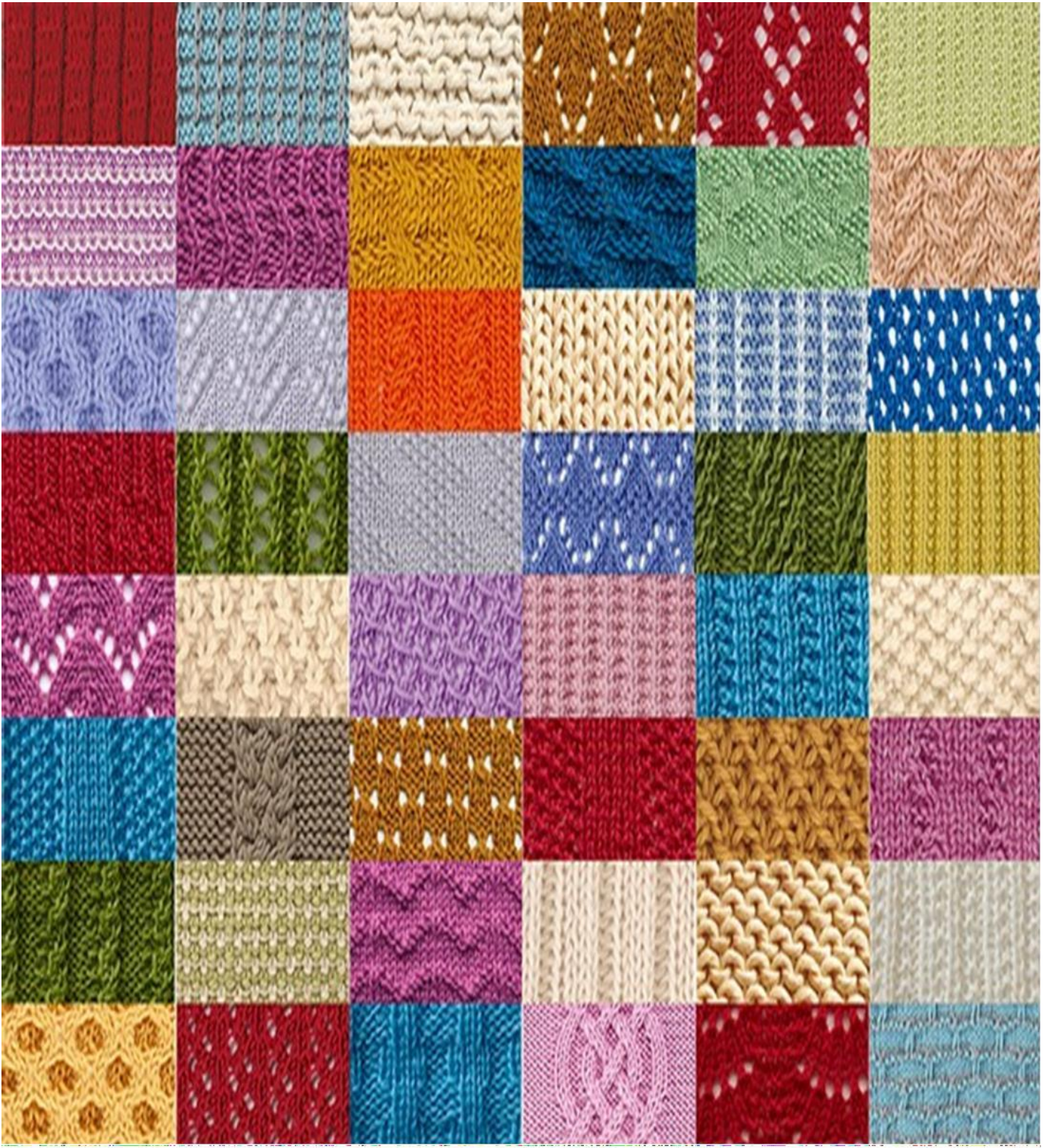

# Using AI & DS for the Automation of Knitting Patterns

**BY: UDUAK UBOH**

## Abstract

Knitting patterns are a crucial component in the creation and design of knitted materials. Traditionally, these patterns were taught informally, but thanks to advancements in technology, anyone interested in knitting can use the patterns as a guide to start knitting. Perhaps because knitting is mostly a hobby, with the exception of industrial manufacturing utilising specialised knitting machines, the use of AI in knitting is less widespread than its application in other fields. However, it is important to determine whether knitted pattern classification using an automated system is viable. In order to recognise and classify knitting patterns. Using data augmentation and a transfer learning technique, this study proposes a deep learning model. The Inception ResNet-V2 is the main feature extraction and classification algorithm used in the model. Metrics like accuracy, logarithmic loss, F1-score, precision, and recall score were used to evaluate the model. The model evaluation's findings demonstrate high model accuracy, precision, recall, and F1 score. In addition, the AUC score for majority of the classes was in the range (0.7-0.9). A comparative analysis was done using other pretrained models and a ResNet-50 model with transfer learning and the proposed model evaluation results surpassed all others. The major limitation for this project is time, as with more time, there might have been better accuracy over a larger number of epochs.

## Introduction & Background

By way of wool traders, knitting is thought to have come to Europe in the 5th century from the Middle East. The Highlands and Scotland quickly developed a native knitting industry, and the knitted goods were exported to the rest of Europe and by the 20th century, knitting had become fashionable (The Sustainable Fashion Collective, 1992). Knitting patterns, as we now refer to instructions on how to construct knit products, have been known by a variety of names over the years and have evolved from being shared informally to being published online to users all over the world.

The similarities between knitting and coding have been increasingly discussed among programmers. Knitting is all about iteratively repeating combinations of knit and purl stitches to make an output. To generate and produce patterns, a symbolic language is used – this is very similar to programming and programming languages. Considering these similarities, it becomes an interesting topic to understand how artificial intelligence can be employed to produce knit patterns.

The modern economy is being altered by new manufacturing techniques that allow for the fully automated fabrication of custom items and parts, and the textile sector is not exempt from this change. Patterns are made up of repeating forms and geometries and are associated with textiles because knitting was the origin of most pattern styles (Stewart, 2015). These visual patterns can be discovered in various forms in nature, technology, mathematics and especially in textiles.

Industrial knitting allows for complete customisation of knitted clothing without stitching - every loop of yarn is electronically controlled by a basic programme to create the completed item, which largely lacks any distinctive knit pattern. The significance of patterning in knitting cannot be emphasised (Shida et al., 2017), yet these pattern libraries (which provide instructions for hundreds of beautiful manual knitting designs) are often not ideal for industrial knitting machines because they are designed for hand knitting.

The Massachusetts Institute of Technology (MIT) researchers have recently created some highly regarded work. The paper develops a method for translating a single knitted pattern image into machine code that specifies the type of stitch to be utilised at each stage of the garment production process using CNNs and a generative adversarial network (GAN) (Kaspar et al.,2019). While this was the initial goal of this research project, it became increasingly clear that timing would be a key limitation for this. Hence, like many of the work done in this area, knitting pattern recognition is the focus of this research. This study will specifically use deep learning models related to data augmentation and transfer learning approach to recognise knit patterns in images.

Similar to knitting pattern recognition, much research has been done in the field of weave pattern identification. To extract features and classify fleeced fabric with defects, an analysis-based dimensionality reduction method, Naive Bayes and K-nearest Neighbor classifiers was used, with an emphasis on straightforward rather than elaborate texture patterns (Yildiz, 2016). An efficient method based on

Transform Invariant Low-rank Textures (TILT) and HOG was presented to recognise woven cloth patterns (Xiao et al., 2018). To extract features for automatic woven fabric detection, Li et al. (2012) employed a novel method. The high dimensional feature data was first reduced using principal component analysis, followed by the use of a support vector machine as a classifier to identify and categorise the woven fabric. The researchers first computed the fabric features using the local binary pattern approach and the grey level co-occurrence matrix. Another study suggested the use of a seven-characteristic method called Local Feature Similarity (LFS) to recognise weave patterns automatically (Guo et al., 2018).

This project follows the structure below:

- The first section of the study presents the research subject, emphasises its goal and evaluates earlier studies on the subject and on similar subjects.

- The second section introduces and justifies the methodology that was used to answer and solve the research subject.

- The experimental framework is presented in Chapter 3. Their interpretation & discussion will be related to how the research goals are addressed.

- To wrap up this project, important findings/ results from the research are presented in Chapter 4 along with practice-oriented/ practical suggestions. It also identifies further areas for research and emphasises the project's shortcomings.

## Methodology

Convolutional Neural Networks (CNNs) have been known to exhibit excellent performance and accuracy when recognising objects, patterns within images and particularly in image processing. Convolution, pooling, and fully connected layers make up a typical CNN and unlike traditional machine learning algorithms, which rely on manually created features, CNN designs automatically learn the smallest descriptive features. In contrast, deeper CNN architectures like ResNet, contain intricate alternating connections between layers; AlexNet and VGGNet are examples of shallow CNN networks which are built by stacking a number of blocks together. Let's look at the VGGNet architecture, ResNet-50 and Inception ResNet-V2 as these are the convoluted neural networks that would be employed in this project.

### VGGNet

A stack of smaller convolutional kernels and 144 million parameters make up the VGGNet architecture, consisting of five max-pooling layers, three fully connected layers, sixteen convolutional layers with smaller kernel size and an output layer with softmax non-linear activation. VGGNet's computing costs are higher since it has more parameters than AlexaNet and uses more memory.

### ResNet

Undoubtedly the most revolutionary development in the computer vision/deep learning world in recent years was Deep Residual Network, which was created by He et al. in 2015. ResNet enables training with hundreds or even thousands of layers while maintaining impressive performance. ResNet's major advantage is that it addresses the issue of vanishing gradient and declining accuracy by introducing identity shortcut connections, making it adaptable, task-dependent, and capable of training incredibly deep neural networks.

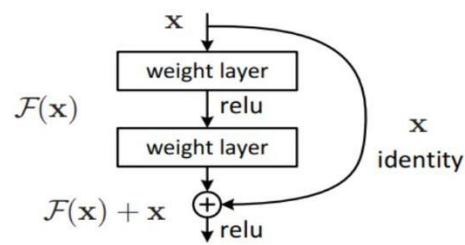

Figure 1: Skip Connections on ResNet

In the figure above, a shortcut is added to the model path. Thus, a very deep network is created by stacking these ResNet blocks on top of one another.

By including skip connections, a model's gradient can be directly backpropagated to prior layers. These shortcut/ skip connections also allow a ResNet block to learn an identity function reasonably quickly. Two main types of blocks are used in a ResNet – the identity block and the convolutional block.

The identity block, which is the default block used in ResNets, corresponds to the situation where the input and output have the same dimension. In the identity block, there is a main path and a shortcut path. A Conv2D with filters (F1) of shape (1,1) and stride (1,1) with valid padding makes up the initial component of the main path, Batch Normalization normalises the channel axis and ReLU activation function is added. The second component is identical to the first component with 'same' padding, whereas the third component is identical to the first component, save for the absence of an activation function. In the final step, the shortcut and input are added and ReLU activation function is applied. See illustration in figure 2 below.

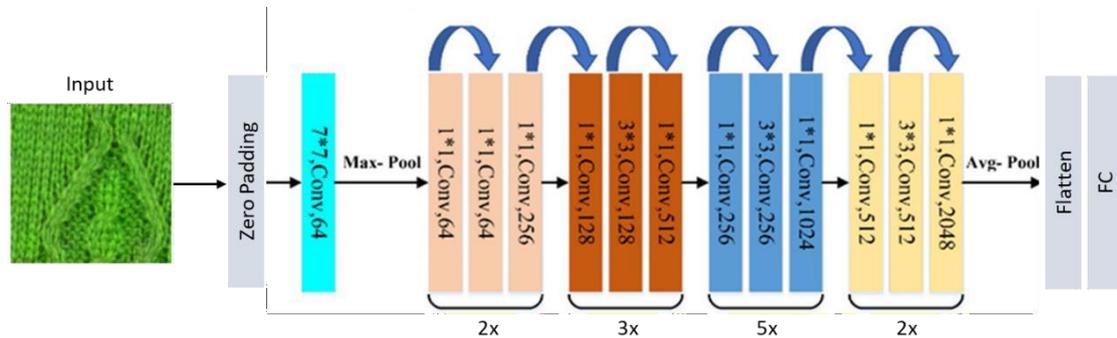

Figure 2: ResNet-50 architecture

Inception ResNet-V2

Inception-ResNet-v2 network is a 164-layers network that is used in image classification as it can accurately classify images into multiple object categories. It builds on the inception architecture but incorporates residual connections (www.mathworks.com, n.d.). Batch normalisation was used in its development to speed up training and using only 7% training steps, Inception-ResNet-v2 employs factorization to minimise the filter size, which minimises the overfitting issue and minimises the number of parameters. Figure 3 below, illustrates the primary structure of Inception-ResNet-v2 which splits the work function into a block consisting of Stem, Inception-ResNet, and Reduction blocks (Gonwirat and Surinta, 2020).

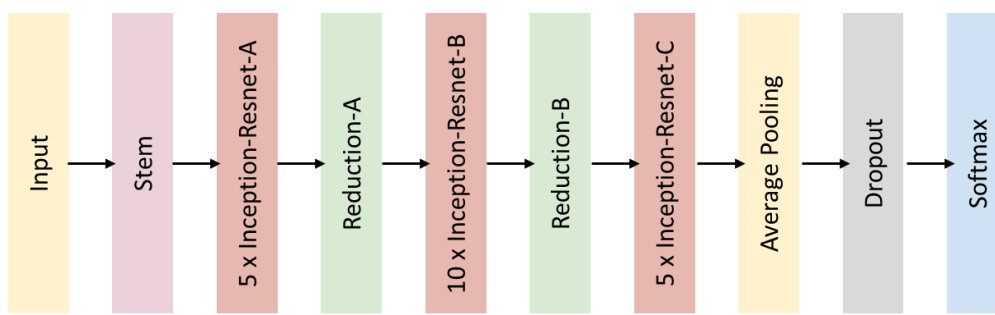

Figure 3: Schema for Inception-ResNet-v2 network

The stem block, which is the first layer, comes before the inception-resnet block. With (3x3) conv filter and stride of 2, the feature map reduces in size, thus, directly lowering the parameters. The next block consists of both inception and ResNet layers. Due to parameter reduction from the stem block, reduction blocks are used to fill the space between the Inception-ResNet block. The final block is the reduction block with a primary role of reducing the size of feature maps within the Inception-Resnet block. There are 2 reduction blocks - one with a single (3x3) conv filter and the other with two (3x3) conv filter & (1x1) conv filter.

Convolutional algorithms are used by CNN to identify patterns within any image in its input. This project's CNN model was trained using a sizable ImageNet dataset. It picks up simple patterns like dots, lines, edges, and diagonals in the first few layers, and then later layers combine these simple patterns to create more complex designs. The final layer can therefore learn important objects. We apply the features (weights) learnt from these layers to the new model through transfer learning. Images of different knit patterns are classified using the key features identified by the pretrained model for the recognition of ImageNet objects. The training procedure is accelerated by the transfer learning technique, and the new CNN model can be easily built as a result.

Project Model

The Inception-ResNet-v2 CNN architecture is the primary model used in this project. As was already noted, Inception-ResNet-v2 has been used to train a wide range of image classification and can correctly classify about 1000 objects in images. From dataset acquisition, this project will proceed through the following stages:

- **Image Pre-processing:** The dataset applied requires pre-processing, image conversions, and scaling. Since there weren't as many pictures of knit-patterns, a variety of augmentation techniques were used to increase the dataset, reducing overfitting and improving recognition.
- **Pattern Scale Identification:** The scale or size of the pictures obtained must be taken into consideration because the project's goal is to recognise knit patterns from images. The input knit pattern image for this project is scaled down before being applied to the model.
- **Model Development and Training:** The inception resnet-v2 model employed in this project optimised the hyperparameters during training.
- **Model Evaluation:** Several metrics, including accuracy, precision, recall, F1-score, and the cross-validation method, were used to evaluate the model's performance. A number of pretrained models were compared with the project model as well.

Dataset

Researchers from the Massachusetts Institute of Technology (MIT) created the dataset that was used. It has 2,091 coloured, 160x160 pixel real photographs of knit patterns and 10,965 synthetic images of knit pattern. An additional dataset gotten from web scraping of knitting websites containing 179 images was also included bringing the total dataset size to 13,235 images. It was pre-processed using image data generator and then split into training, validation and test dataset using ratio 70:20:10. The dataset contains images of 7 types of knit stitch patterns and represent 7 classes – Knit & Purl Stitch, Cable Stitch, Diamond Stitch, Moss Stitch, Mesh Stitch, 'Motif Stitch & Tuck Stitch. These classes were imbalanced, thus class weights were calculated and applied only to the primary Inception-ResNet-V2 model.

Data Augmentation

The quantity and variety of data available during training heavily influences how accurately the Supervised Deep Learning models make predictions. The dataset already contains a significant number of synthetic images which helped to increase the original data size. However, to synthesize more data, several transformation techniques were applied to the original dataset and employed during training, to not only increase the training data size but also to mitigate against overfitting. The image pixels are rescaled to from 0-255 RGB coefficients to 0-1, which helps to normalise the input data. In this project, the images were randomly flipped horizontally, randomly rotated by 20 degrees, images were randomly zoomed in by 80% and zoomed out by 120%. A shear range of 0.2 was used to distort the image along an axis of 20% to create perception angles for the model. Empty spaces from the rotation technique were filled with the nearest pixel value. In figure 4 below, an example of the augmented photos is displayed.

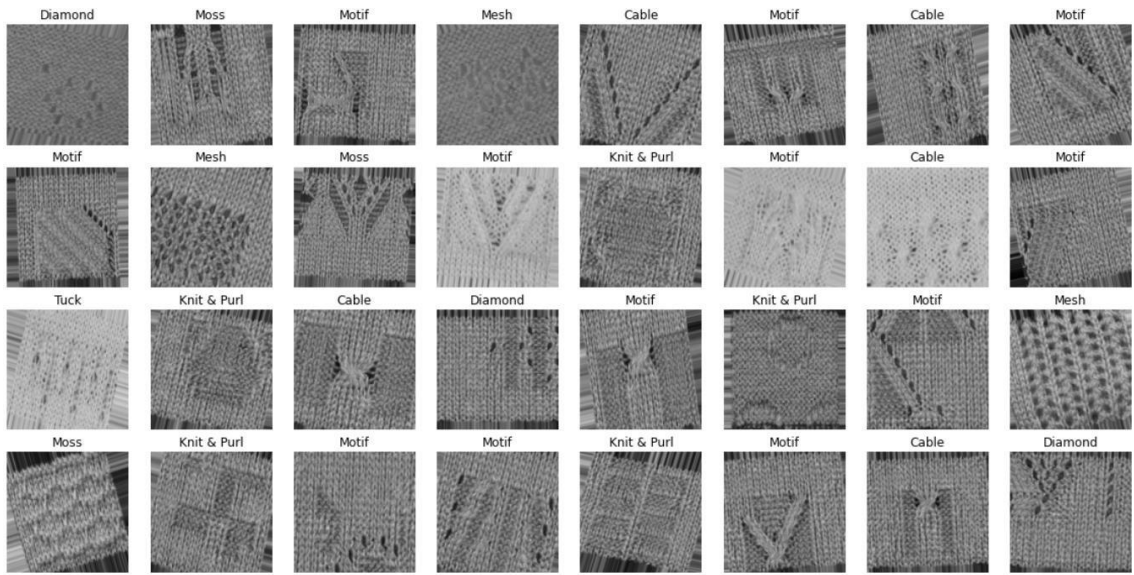

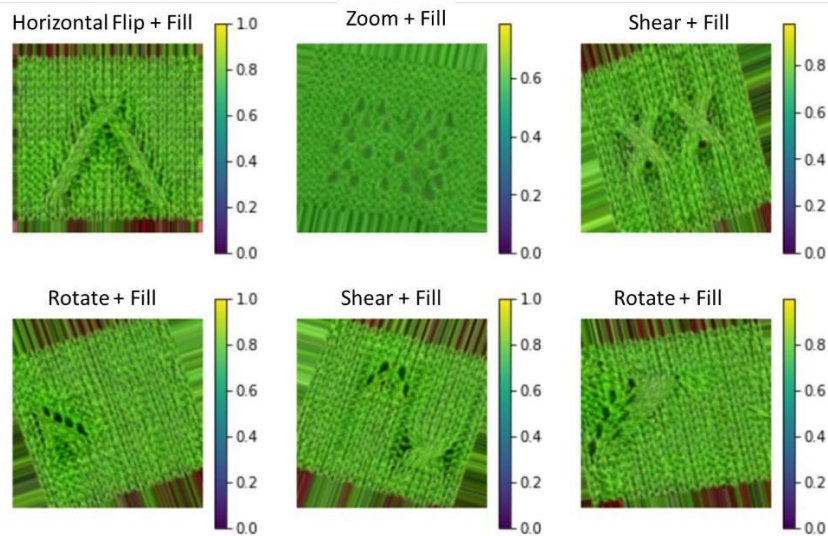

Figure 4: Augmentation Techniques Applied to Original Image

# Experimental Framework

In addition to being resized to 224 by 224 dimensions, the photos of knit patterns were also pre-processed by converting all images from RGB to grayscale to simplify the algorithm and reduce computational requirements. 6 CNN architectures using pretrained model, imagenet weights, adam optimizer with a learning rate of 1e-5 was applied to the dataset to provide a comparison. These models were downloaded directly from the Keras API and used to classify the model. The aim was to determine whether the pretrained models achieved higher accuracy than the new model. Using the weights learned from ImageNet, the transfer learning method was applied to the Inception-ResNet-v2 pretrained CNN architecture. To facilitate feature extraction, the option include top=False was added which eliminated the final dense layers; as a result, the model's input and output can be controlled. A global average pooling, which connects the dimensions of the previous layer with the new layers, was used to connect the base/pretrained model with new layers of the new model. Additionally, three pairs of fully connected layers were then stacked. A ReLU activation layer and a dropout of 0.5 was incorporated in the fully connected layers, which also had 512, 256, and 128 neurons, respectively. The new model incorporates optimizers to produce results more quickly and regularizers to prevent overfitting. The crucial parameters taken into account are the below, each of which can have an impact on accuracy:

- Batch size: Because they are compatible with the computer's memory, it is advised to employ a variety of batch sizes with powers of 2 (8, 16, 32, 64, 128, etc.). A batch size of 64 was used.
- Number of layers: Additional layers were needed for the new model as feature extraction was incorporated.
- Optimization methods: Adam, SGD, RMSprop and Nadam were used to test the new model. SGD & RMSprop had a very low learning capacity and needed a significant number of epochs to run. Adam with 25 epochs was therefore used.
- Learning rate: To avoid significantly altering prior knowledge, a relatively low learning rate is advised for transfer learning. The default learning rate for Adam optimizer (0.001) was applied.
- Regularization methods: Batch normalisation and dropout were applied between the dense layers to avoid overfitting.
- Call-backs: To reduce overfitting without sacrificing model accuracy, an early stopping call back with patience of 10 and a minimum delta of 0.001 was incorporated into the new model.

The new model was trained on "NVIDIA Tesla K80" GPU, using google Colab environment with 26 GB RAM. Python 3.7.13 is employed throughout this project with Keras & TensorFlow libraries for model implementation.

## Results & Discussion

Model evaluation is a crucial component of any machine learning project. There are different types of evaluation metric and the most popular is the accuracy score.

However, accuracy alone is insufficient to determine model efficiency as a model could have a high accuracy score and perform terribly when using an evaluation metric such as logarithmic loss. This project's model will be evaluated using accuracy, logarithmic loss, confusion matrix, precision, recall & F1 score. Logarithmic loss (log loss) is a classification loss function that measures the cost associated with inaccurate predictions in classification tasks. For this model, categorical cross entropy was employed because the task is a multiclass classification. The accuracy and loss for this model is 77% & 0.91 respectively. While this is a relatively good result, it is not enough to evaluate based on this factor alone. The correlation between the expected and actual classes is shown in a confusion matrix. Each row of the matrix represents the predicted class, while each column represents the actual class. When models can identify which classes are most susceptible to confusion and take into consideration the overlap in their respective class attributes, they are considered to be successful. Precision measures the proportion of correct instances among the retrieved instances while recall which relates to the relevant instances that were retrieved. The precision and recall scores are on average 16%. The F1 Score, which ranges from [0, 1], is the harmonic mean of precision and recall. F1 not only shows how precise a model is but also how robust it is. The average F1 score for the project is 18%. See the illustration below:

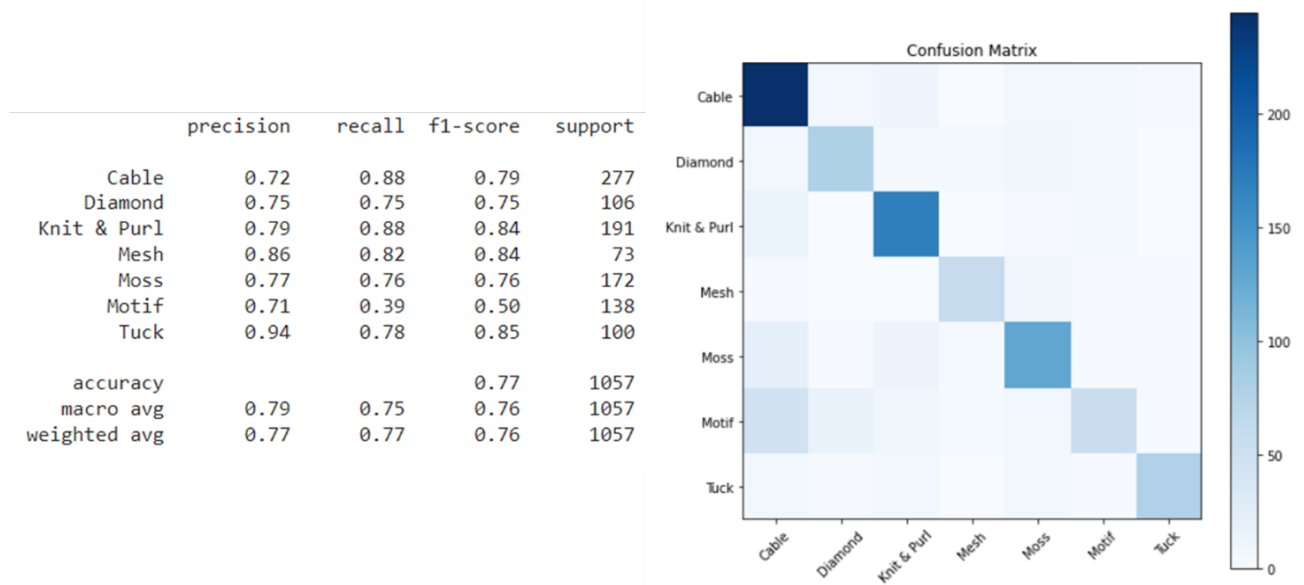

Figure 5: Confusion Matrix and Classification Report for Inception Resnet-V2 Model

A performance measure for the multiclass classification at different threshold values is the roc-auc curve. AUC stands for the level or measurement of separability, while

ROC is a probability curve. It indicates how well the model can distinguish between classes. The more effective the model is in classifying, the higher the AUC. The roc-auc score for the model is 0.8029.

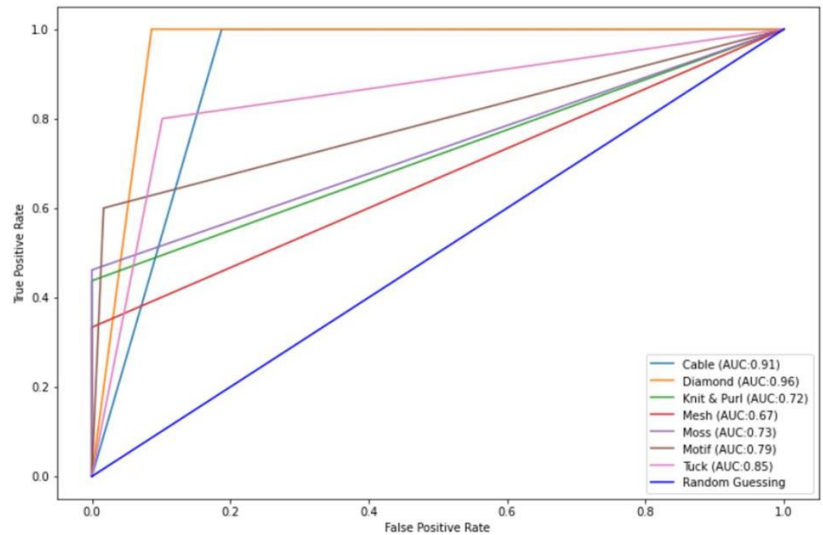

Figure 6: ROC-AUC Curve for Inception Resnet-V2 Model

A comparison analysis was carried out using pretrained models - VGG19, InceptionV3, MobileNetV2, ResNet101, ResNet152, ResNet50. These models were trained using 2 epochs and with imagenet weights. In addition, ResNet-50 pretrained model was applied as a base network and applied to a new model using transfer learning. This model trained was trained using the same parameters as the primary Inception-Resnet model.

| Model | Train Loss | Train Accuracy | Test Loss | Test Accuracy |
|---|---|---|---|---|
| InceptionV3 | 1.0731 | 0.6104 | 0.9954 | 0.6399 |
| MobilenetV2 | 1.0906 | 0.6088 | 1.3256 | 0.5146 |
| Vgg19 | 1.5894 | 0.3810 | 1.4872 | 0.4524 |
| Resnet-101 | 1.7848 | 0.2983 | 1.7436 | 0.2970 |
| Resnet-50 | 1.8398 | 0.2686 | 1.8081 | 0.2869 |
| Resnet-152 | 1.8475 | 0.2641 | 1.8368 | 0.2635 |
| Resnet-50 (trf-learn) | 1.0392 | 0.6119 | 1.0423 | 0.6108 |
| Inception-Resnet-V2 | 0.1462 | 0.9499 | 0.6273 | 0.8498 |

Overall, the model evaluation results above show that the model was sufficiently able to classify knit patterns into the different classes. Worthy of note is that due to the imbalanced dataset, it was imperative to include class weights into the model so as to achieve a good performance. Thus, while the base model learned from the pretrained imagenet weights, the proposed model employed distinct class weights to reduce bias in classification.

Limitations

This study has significant and wide-ranging limitations. Every effort was made to find alternatives, but occasionally I was forced to accept the limitations. One of the main restrictions was time. While my proposed model performed fairly well, more time would have allowed me to train the model with up to 200 epochs to increase its accuracy steadily over time. Unfortunately, Inception-ResNet models are known to train over a long period of time and with trying to fit the model using various optimizers and learning rates and different epochs just to increase the accuracy and reduce overfitting, there was not much time to train the final model.

Ethical Considerations

An ethical approval was obtained from the Faculty of Science and Engineering in full compliance with university guidelines. All research-related information used in this study was already in the public domain and have been properly cited.

Recommendations for Further Research

In comparison to woven fabrics, the application of AI & data science in knitting are lower and less popular. This may be because knitting is typically done as a hobby and because the faster speed and design of circular knitting machines impose a limitation when applying AI in industrial or machine manufacturing of knitted clothes.

Research and development in AI could include monitoring & validation of defects in knitting fabrics as they run through the machines at high speeds. Furthermore, deeper research into the use of AI in 3D pattern prediction by creating a knit software that generates knitting patterns and print these patterns using 3D.

# Conclusions

The goal of this research was to develop a customized deep learning model for knitting pattern recognition. Specifically, this project is based on the Inception-ResNet-V2 model architecture and leveraged deep learning models relating to data augmentation and transfer learning approach to recognise knit patterns in images.

Image Pre-processing was executed – the image dataset was resized to an appropriate scale and normalised, RGB colour was changed to grayscale to simplify the algorithm and reduce computational requirements. Several augmentation approaches were executed to increase the dataset size including shearing, flip and zooming. A pretrained Inception-Resnet-v2 model is employed as base model and all layers except the last block is frozen. This is then added to 3 pairs of fully connected, dropout and batch normalisation layers. The features in the knit patterns are extracted and then classified into their respective classes - Knit & Purl Stitch, Cable Stitch, Diamond Stitch, Moss Stitch, Mesh Stitch, Motif Stitch & Tuck Stitch.

The model's performance was evaluated using a variety of evaluation metrics including accuracy, logarithmic loss, precision, recall, F1-score & ROC-AUC curve and it produced promising results with its only limitation being training time.

Finally, recommendations for future research were provided in the hopes that more research can be done in applying AI to 3D knit pattern generation and the monitoring & validation of defects in knitting fabrics as they run through the industrial circular knitting machines at high speeds.